\documentclass{isprs} 
\usepackage{subfigure}
\usepackage{setspace}
\usepackage{geometry} 
\usepackage{epstopdf}
\usepackage[labelsep=period]{caption}  
\usepackage[british]{babel} 
\usepackage[hang]{footmisc}
\usepackage{natbib}
\usepackage{amsmath}
\usepackage{multirow} 
\usepackage{booktabs}
\usepackage{graphicx}
\usepackage{adjustbox}
\usepackage{amssymb}

\begin{document}

\title{Semantic Segmentation of Textured Non-manifold 3D Meshes using Transformers}
\date{}


\author{Mohammadreza Heidarianbaei, Max Mehltretter, Franz Rottensteiner}

\address{
Institute of Photogrammetry and GeoInformation, 
Leibniz University Hannover, Germany\\
(heidarianbaei, mehltretter, rottensteiner)@ipi.uni-hannover.de

}



\abstract{

Textured 3D meshes jointly represent geometry, topology, and appearance, yet their irregular structure poses significant challenges for deep-learning-based semantic segmentation. 
While a few recent methods operate directly on meshes without imposing geometric constraints, they typically overlook the rich textural information also provided by such meshes. We introduce a texture-aware transformer that learns directly from raw pixels associated with each mesh face, coupled with a new hierarchical learning scheme for multi-scale feature aggregation. 
A texture branch summarizes all face-level pixels into a learnable token, which is fused with geometrical descriptors and processed by a stack of Two-Stage Transformer Blocks (TSTB), which allow for both a local and a global  information flow. 
We evaluate our model on the Semantic Urban Meshes (SUM) benchmark and a newly curated cultural-heritage dataset comprising textured roof tiles with triangle-level  annotations for damage types. 
Our method achieves 81.9\% mF1 and 94.3\% OA on SUM and 49.7\% mF1 and 72.8\% OA on the new dataset, substantially outperforming existing approaches. 

}


\maketitle



\section{Introduction}\label{MANUSCRIPT}
 
Understanding 3D structure is a fundamental task in various domains,  e.g.\ 
Geospatial Analysis \citep{KOLLE2021100001} and Cultural Heritage (CH) preservation  \citep{yang2023three}. 
Textured 3D meshes offer unique advantages over alternative 3D surface representations such as point clouds or voxel grids: 
Such meshes represent the geometry and topology of 3D objects by connecting vertices by edges and faces. 
Additionally, they contain textural information, with  every face being linked to a corresponding area in a 2D image (texture map), enabling the integration of surface appearance attributes such as colour, pattern, and material properties at a higher spatial resolution than the point cloud corresponding to its vertices.


However, due to the irregular structure of 3D textured meshes, their semantic segmentation  remains a challenge for deep learning methods. 
Existing methods for processing meshes \citep{feng2019meshnet, hu2022subdivision,milano2020primal}  still exhibit limitations. 
A major problem is the reliance on topological constraints (e.g.\ manifoldness) that are often violated in real-world meshes, which  restricts the applicability of such methods. 
While some works attempt to handle non-manifold meshes, e.g.\ \citep{sharp2022diffusionnet, heidarianbaei2025nomeformer}, most of them focus on geometrical features, considering texture in simplistic ways only. 
Some methods rely on low-dimension\-al descriptors, e.g.\  a single RGB vector for every vertex  \citep{zi2024urbansegnet}. 
Such a representation is spatially sparse and fails to capture  high-frequency details available in the texture map. 
Other approaches resort to statistical descriptors to represent  texture, e.g.\  mean colour vectors or histograms. 
While such descriptors provide a compact representation, they  disregard spatial patterns in the texture that may be crucial for distinguishing  different object types. 
This difficulty is partly due to the fact that the number of texture pixels associated with each face can vary significantly, making the direct application of conventional  architectures, e.g., convolutional neural networks, non-trivial.


Recent  work on semantic segmentation has increasingly relied on  transformer-based architectures based on the attention mechanism \citep{vaswani2017attention} due to the ability of such networks to consider global context. 
However, their computational complexity often renders full attention impractical, leading to the requirement for patchification, i.e.\ a reduction of the spatial resolution of the input data. 
In the domain of 3D mesh processing, \citet{heidarianbaei2025nomeformer} presented a transformer-based approach that avoids a reduction of the spatial resolution, addressing this problem by partitioning mesh faces into clusters and applying self-attention within each cluster to capture local context while also determining one  token encoding the entire  cluster ({\em cluster token}). 
The cluster tokens interact with each other and with all face tokens via cross-attention to model global dependencies.
We believe that, despite the residual connections used in transformers, this formulation might lead to an over-smoothing of the classification results, as face-level tokens are updated by weighted averages of cluster tokens. 


In this paper, we propose a new method for semantic segmentation of textured 3D meshes. 
It builds on \citep{heidarianbaei2025nomeformer} due to its capacity for processing non-mani\-fold meshes, but we extend it to address the limitations of existing methods   mentioned earlier. 
On the one hand, we introduce a transformer-based texture branch that directly processes all pixel values associated with each face and delivers a texture feature vector that is fused with  geometrical features, resulting in a unique representation of every face of the mesh by a vector capturing both cues.
On the other hand, we propose  a modified hierarchical attention mechanism that enhances inter-cluster information exchange while preserving local feature diversity and, thus, reducing potential oversmoothing effects of the original cross-attention-based approach. 
This is achieved by restricting the global information exchange to  cluster tokens via self-attention, after which each cluster token propagates the aggregated context back to the faces of its cluster. 
This  process enables an efficient bidirectional information flow between local and global levels, considering  context at different scales without blurring fine-grained local features.


We evaluate our proposed method using two datasets,  the SUM benchmark dataset for urban classification \citep{gao2021sum} and a new dataset from the domain of Cultural Heritage preservation  consisting of   textured 3D mesh patches of the roof of a historical building with triangle-level annotations for  damages such as {\em biological colonization}, showing the applicability of the method to different domains.  
Our main contributions can be summarized as follows:
\begin{itemize}
    \item We introduce a new unified transformer-based architecture for semantic segmentation of textured 3D meshes that jointly considers geometrical and textural features, deriving the latter  directly from the pixels associated with each mesh face. Unlike prior approaches  relying on coarse colour statistics or vertex-level features, our method explicitly aggregates per-face texture information through a dedicated texture branch.
    \item  We propose a new method for capturing long-range interactions in transformer-based semantic segmentation of 3D meshes  to facilitate effective information exchange between local and global contexts while avoiding over-smoothing and preserving fine-grained geometric detail. 
    \item We evaluate our approach on the two datasets mentioned earlier, showing its applicability in two different application domains (urban classification, cultural heritage preservation). We also report the results of ablation studies validating the effectiveness of our design choices.
\end{itemize}


\section{Related Work} 
\label{sec:Related_Work}

Processing 3D data has witnessed significant advancements in recent years, largely driven by progress in deep learning \citep{zhang2025survey,ioannidou2017deep}. 
Most existing methods focus on a specific representation, e.g. on voxel-based representations of 3D data \citep{maturana2015voxnet, wu20153d, riegler2017octnet} or on point clouds  \citep{qi2017pointnet++, hu2020randla, qian2022pointnext}, which provide a more efficient and explicit way to represent the geometry of 3D objects. 
Despite these advancements, such methods cannot directly operate on 3D meshes  as input. 
Meshes are based on an inherently  irregular data structure, posing significant challenges for neural network architectures originally designed for regular grids. 
As a result, mesh-based semantic segmentation is still underexplored. A common workaround is to convert mesh faces into point clouds, e.g.\ using face centroids with texture descriptors \citep{Laupheimer2022}, and then to apply networks for point cloud segmentation, e.g.\  \citep{qi2017pointnet++}. 
However, it would be desirable to fully exploit the inherent potential of mesh data. 


There are just a few works on the direct classification of triangles in a mesh. 
Early work extended convolutional neural networks to operate directly on 3D meshes by redefining convolution and pooling for non-Euclidean geometry. 
For instance, geodesic CNNs \citep{masci2015geodesic} apply convolution to cur\-ved sur\-faces using geodesic coordinates instead of grid structures.
MeshNet \citep{feng2019meshnet} processes mesh faces using spatial and structural descriptors and expands the receptive field by mesh-based convolution layers. 
\citet{hu2022subdivision} focus on hierarchical feature learning, including subdi\-vision-based  pooling to build fine-to-coarse mesh representations. 
\citet{hanocka2019meshcnn} consider edges to be the basic elements and use an edge-based pooling strategy to adapt mesh topology and capture multi-scale structure.
Another research direction leverages the graph structure of meshes. 
\citet{milano2020primal} extend point-wise convolutions to meshes by constructing face- and edge-based graphs and aggregating  features via graph attention networks, using mesh simplification  for pooling. 
MeshWalker \citep{lahav2020meshwalker} performs random walks on the vertex graph and encodes the resulting sequences with recurrent networks to capture local topology.
However, all of these  methods require manifold mesh structures, enforcing well-defined neighbourhood relationships that enable the adaptation of conventional deep learning operations. 
In practice, 3D meshes produced in automated photogrammetric pipelines often violate these manifold constraints, limiting the applicability of such approaches to real-world data \citep{KOLLE2021100001,gao2021sum}.


There is some  work that relaxes manifold assumptions to better handle irregular or non-manifold geometry. 
Laplacian2Mesh \citep{dong2023laplacian2mesh} operates in the spectral domain by leveraging the mesh Laplacian, enabling processing of non-manifold structures but focusing primarily on vertex-level tasks and risking loss of fine spatial detail. 
DiffusionNet \citep{sharp2022diffusionnet} performs directional filtering using pointwise perceptrons with learned diffusion and spatial gradients, though aggregation remains limited to local neighbourhoods. 
In contrast, \citet{tutzauer2019semantic} extract multi-scale face descriptors and apply a 1D CNN, with spherical neighbourhoods providing geometric context; the convolution  is mainly used for feature embedding and not for true spatial aggregation. 
None of the  methods cited so far incorporates textural information, and their focus remains primarily on geometric and object-level data.


MFSM-Net \citep{hao2025mfsm} represents texture by a single orthophoto derived from the mesh. 
Textural and geometrical features are extracted from the orthophoto and the mesh in two separate streams before being fused. 
However, a single image cannot fully represent the radiometric information of all mesh faces in real 3D scenarios.  
More recently and most relevant to our method, \citet{heidarianbaei2025nomeformer} introduced a transformer-based architecture for the per-face semantic segmentation of textured non-manifold meshes. 
They construct compact face descriptors by combining handcrafted geometrical features with simple texture statistics. 
To cope with the irregular mesh structure and the large number of faces, the faces are spatially clustered. 
A Local–Global (L-G) transformer architecture captures both local and global context: local self-attention operates within clusters with the support of  cluster tokens, while a global cross-attention mechanism enables information flow across clusters at substantially reduced computational cost. 
However, that method does not directly process the raw texture information associated with each face, but  relies on few statistical descriptors that are not expected to capture fine-grained textural patterns. 
Second, the cross-attention mechanism in the L-G blocks updates the per-face features as weighted averages of cluster tokens, which we believe to lead to oversmoothing of the classification results.


To address the limitations of the cited approaches, we propose a new network architecture for semantic segmentation of 3D meshes. 
The new method extends \citep{heidarianbaei2025nomeformer}  by a texture-embedding module and an improved transformer-based mechanism for considering global context. 
To the best of our knowledge, no existing deep learning framework directly integrates raw texture information  alongside geometrical features at triangle level for the task at hand; the exception  \citep{hao2025mfsm} requires a projection to a common plane, which might be problematic in real 3D scenarios. 
We also believe that the proposed attention-based mechanism will allow the model to propagate high-level contextual knowledge back to individual faces with\-out overriding the local information completely, thus allowing to predict class labels correctly at a fine local scale.


\section{Methodology}

\subsection{Overview}
\label{Overview}

The input to our method consists of a textured 3D triangulated mesh
\(
M = \left( V, F, T, I \right),
\)
where \(V\) denotes the set of vertices, each corresponding to a point in 3D space, and \(F\) denotes the set of triangular faces. 
Each face is defined by an ordered list of three vertices. 
The texture is contained in a texture image \(I\); \(T\) denotes the information required for linking every face to an area in~\(I\), i.e.\ the  coordinates of the points in \(I\) corresponding to the three vertices of the face ({\em texture coordinates}).
The primary objective of the network is to perform semantic segmentation. 
The output is represented by 
\(
\widehat{Y} \in \mathbb{R}^{|F| \times C},
\)
where \(|F|\) is the number of faces and \(C\) is the number of predefined classes. 
\(\widehat{Y}\) contains a vector of \(C\) class scores for every face of the mesh. 
The class label of a face is determined as the class having the maximum score. 

\begin{figure*} 
    \centering
    \includegraphics[width=0.8\textwidth]{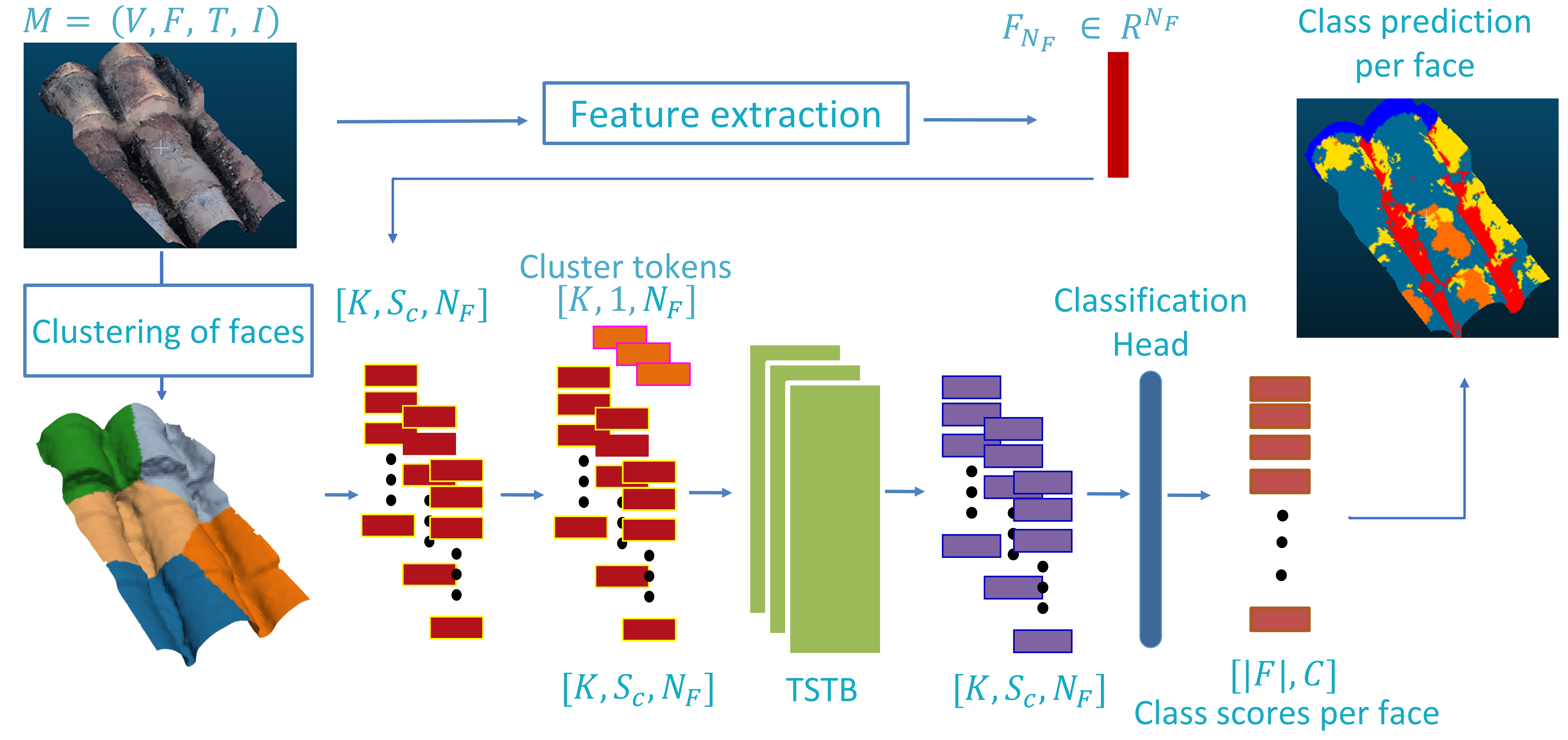}
    \caption{Overview of the network architecture. The feature extraction branch generates one feature vector per face. The faces are clustered using K-means clustering. The tensor containing the feature vectors is reshaped so that there is one sequence of feature vectors per cluster, and a learnable cluster token is prepended to each sequence. These feature vectors are  processed by a sequence of $N_{Bl}$ TSTB. The output of the final block is used to predict the class scores and  labels. Numbers in brackets denote tensor dimensions.}
    \label{fig:architecture}
\end{figure*}

As shown in Figure~\ref{fig:architecture}, the input mesh is first processed by a feature extraction branch, which generates a feature vector $F_{N_F}$ of dimension $N_F$ for every triangular face in the mesh. 
This vector encodes both geometrical and textural information related to a face. 
However, unlike most existing work, our network directly learns to determine representations from the raw pixel data using a transformer-based network. 
The resulting per-face feature vectors are aggregated into a tensor of size $|F| \times N_F$, which is then passed to the subsequent components of the network.
The feature extraction branch is described in Section~\ref{Feature_Extraction}.


The core  module of our network builds upon transformer architectures \citep{vaswani2017attention}. 
However, the computational cost of standard self-attention scales quadratically with the number of faces~$|F|$, which becomes prohibitive for meshes containing a large number of faces. 
To overcome this limitation, inspired by prior works \citep{chu2021twins,heidarianbaei2025nomeformer}, we adopt a hybrid local–global processing strategy and propose the so-called Two-Stage Transformer Block (TSTB), in which face feature vectors are processed in two successive stages operating on a local and a global level, respectively.


Following  \citep{heidarianbaei2025nomeformer}, we first construct a spatial partition of the mesh: the vertices~$V$ are clustered into~$K$ subsets using $K$-means clustering \citep{Bishop2006}. 
Each face is assigned to the cluster that contains the majority of its vertices; if all three vertices lie in different clusters, the face will be assigned randomly. 
The input feature tensor is reshaped so that there is one sequence of feature vectors ({\em tokens}) per cluster, and a single learnable cluster token is prepended to each sequence. 
This results in a tensor $\mathbf{X}$ of dimension $K \times (S_c+1) \times N_F$, where $S_c$ denotes the maximum number of faces contained in any cluster.  
To enable efficient implementation and parallel processing across clusters, all clusters with fewer than $S_c$ faces are zero-padded to this size; a binary attention mask is generated to identify the padded feature vectors, so  that attention will only be computed  among valid (non-padded) tokens. 
In the first stage of a TSTB, self-attention is restricted to operate within the clusters. 
This localized attention reduces the computational footprint while preserving fine-scale cues. 
To enable long-range information exchange, we propose to use a self-attention mechanism among cluster tokens in the second stage of a TSTB, which enriches them with long-range contextual information. 
The enriched cluster tokens will interact with their local face tokens in the first stage of the following block, enhancing global–local consistency while preserving the discriminative details of individual face vectors. 

The last TSTB delivers the set of face representations that serves as the input of  the classification head, which discards cluster tokens and predicts the class scores and labels. 
The TSTB and the classification are described in Sections~\ref{sec:Bi-Level} and~\ref{sec:classification_head}, respectively.
Section~\ref{sec:loss} presents the training procedure. 


\subsection{Feature Extraction Branch}
\label{Feature_Extraction}
This branch extracts a feature vector for each face by combining geometrical and textural cues. 
It consists of two independent sub-branches: the {\em geometry branch} producing geometrical features $\mathbf{F}_G \in \mathbb{R}^{d}$, and the {\em texture branch} generating textural features $\mathbf{F}_T \in \mathbb{R}^{d}$. 
The two vectors are concatenated to form $
\mathbf{F} = [\mathbf{F}_G, \mathbf{F}_T] \in \mathbb{R}^{2d}$,
which is processed by a two-layer multi-layer perceptron (MLP) to obtain the final face-specific feature vector  $\mathbf{F}_{\mathrm{out}} \in \mathbb{R}^{N_F}$. 
The MLP uses ReLU activations and dropout for regularization. 
The sub-branches are described below.


\subsubsection{Geometry Branch:}
\label{Geometry Branch}
Similarly to \citep{heidarianbaei2025nomeformer}, we derive 16 hand-crafted features per face that encode  shape and positional cues: normalized vertex coordinates, normal vector components, area, and three  angles with the adjacent faces. 
A linear layer projects the raw descriptor
$\mathbf{F}_G^{\text{raw}} \in \mathbb{R}^{16}$ into the final geometrical embedding
$\mathbf{F}_G \in \mathbb{R}^{d}$.


\subsubsection{Texture Branch:}
\label{Texture Branch}
In order to determine the texture feature vectors, the pixels associated with a face are extracted from the texture image \(I\) inside the area indicated by the texture coordinates and stored in a tensor $\mathbf{F}_T^0$ of dimension $P \times Ch$, where $P$ is the number of pixels in the face (which varies from face to face) and $Ch$ is the number of channels ($Ch = 3$ for RGB texture images). 
As the model requires the same feature dimensionality for each face, the texture branch  determines a texture feature vector $\mathbf{F}_T \in \mathbb{R}^d$ from this information that encodes the texture information of the face.


The texture branch, which is a variant of a transformer network \citep{vaswani2017attention}, takes $\mathbf{F}_T^0$ as input. First, a learnable token (referred to as the \emph{texture token}) is appended to the tensor $\mathbf{F}_T^0$, resulting in an augmented input tensor $\mathbf{F}_T^{\mathrm{raw}}$ of size $(P+1) \times Ch$. 
$\mathbf{F}_T^{\mathrm{raw}}$ is  processed by a fully connected layer that maps the features into a space of dimension $d$,  resulting in a tensor $\mathbf{F}_T^{\mathrm{inp}}$ of dimension $(P+1) \times d$ containing $P$ transformed feature vectors and the transformed texture token. 
This tensor is processed by a transformer block~\citep{vaswani2017attention} using multi-head self-attention. 
The MLP inside the block consists of two fully connected layers with ReLU activation and uses  an intermediate feature dimension of $2d$. 
As the texture token interacts with all pixels of a face,  we only retain  this token after applying the transformer block and consider it to be the final texture feature $\mathbf{F}_T \in \mathbb{R}^{d}$ of the corresponding face.


\subsection{Two-Stage Transformer Blocks (TSTB)}
\label{sec:Bi-Level}
\begin{figure}[ht]
    \centering
     \includegraphics[width=0.45\textwidth]{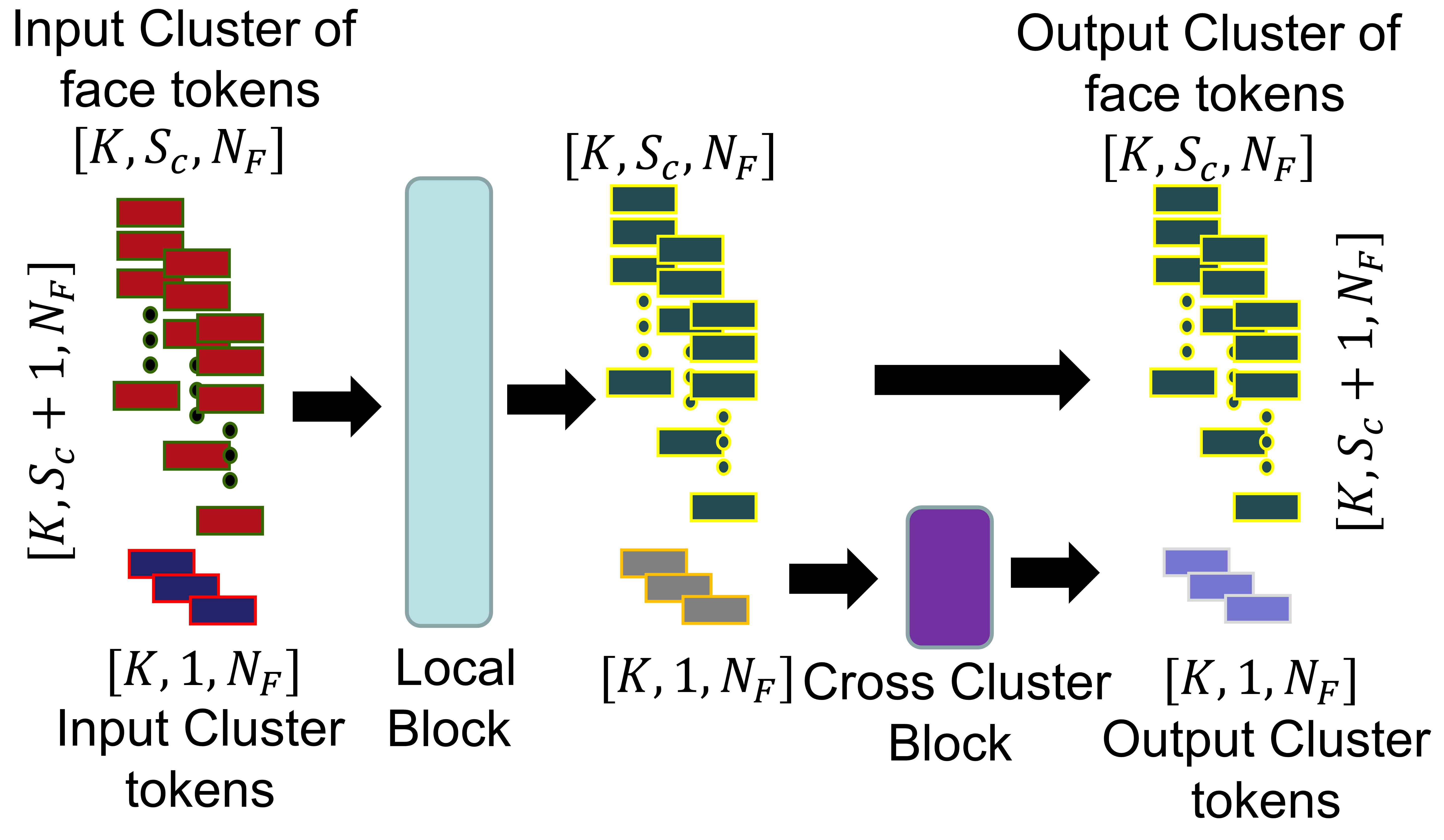}
    \caption{Architecture of a TSTB. It consists of two sub-blocks: a local  block in which the feature vectors within each face cluster can interact with each other, and a cross-cluster block designed to capture global context by allowing the cluster tokens of different clusters to interact. Numbers in brackets are dimensions of  tensors.}
    \label{fig:lg-transformer}
\end{figure}

The TSTB is the main building block of our architecture. 
An overview of the structure of such a block is given in Figure~\ref{fig:lg-transformer}. 
Each block $j$ receives an input in the form of a tensor
$\mathbf{E}_j$  of dimension $K \times (S_c+1) \times N_F$. 
This tensor contains the face-specific feature vectors together with the cluster tokens. 
The TSTB  generates an output tensor $\mathbf{Z}_j$ of the same dimension. 
In each TSTB, the feature vectors are updated as described below. 
After each block, the binary attention mask introduced in Section~\ref{Overview}  is applied to reset the padded feature values to zero. 
The input to the first block ($j=1$) is the tensor $\mathbf{X}$ that is generated from the outputs of the feature extraction branch and the learnable cluster tokens (Section~\ref{Overview}), i.e., $\mathbf{E}_1=\mathbf{X}$.


The structure of the TSTB is designed to capture both local and global context  from the input mesh. 
As discussed in Section~\ref{Overview}, applying attention across all feature vectors simultaneously would be computationally prohibitive. 
Therefore, each TSTB contains two transformer-based sub-blocks: 
\begin{enumerate}
    \item The \textbf{Local Block} processes the feature vectors within each cluster independently, enabling local context aggregation. This sub-block is described in  Section~\ref{sec:localblock}.  
    \item The \textbf{Cross-Cluster Block} allows the cluster tokens to interact across clusters, capturing global dependencies. It is presented Section~\ref{sec:globalblock}. 
\end{enumerate}


Combining  local and cross-cluster blocks provides a hierarchical mechanism for progressively integrating fine-grained geometrical details with global structural context. 
The local block focuses on intra-cluster interactions, enabling each face feature to exchange information with its neighbours and the corresponding cluster token. 
On the other hand, the cross-cluster block operates on the set of cluster tokens, facilitating inter-cluster communication through self-attention between cluster tokens. 
By updating cluster tokens with information from all other clusters, the cross-cluster block considers long-range dependencies and ensures global consistency across the mesh. 
Glo\-bal context flows back to individual faces when the updated cluster tokens are reintegrated into the local blocks of subsequent layers. 
This alternation  between local and global attention enables the model to capture both detailed local relations and long-range structural patterns, which we expect to lead to more expressive representations.


\subsubsection{Local Block: }
\label{sec:localblock}

Each local block is structured as a standard Transformer block~\citep{vaswani2017attention}, denoted by $TB$. 
The input features are linearly projected into query, key, and value matrices, and multi-head self-attention is applied within each cluster of tokens. 
Following the attention operation, the output is normalized and passed through a two-layer MLP with ReLU activations, with residual connections applied to both sub-modules. 
To formalize this process, let the input tensor of block $j$ be $\mathbf{E}_j$. 
For each cluster $k$, the corresponding input slice is defined as
\[
\mathbf{E}_{k,j}
=
\big[\mathbf{F}^{Cl_k}_{N_F}, \mathbf{F}^{k_1}_{N_F}, \ldots, \mathbf{F}^{N_k}_{N_F}, \ldots, \mathbf{F}^{S_c}_{N_F}\big]_j
\in \mathbb{R}^{(S_c+1)\times N_F},
\]
where the first vector, $\mathbf{F}^{Cl_k}_{N_F}$, corresponds to the cluster token and $\mathbf{F}^{k_n}_{N_F}$ is the feature vector of the $n^{th}$ face in cluster $k$; the feature vectors with a face index larger than the number $N_k$ of faces in that cluster are zero-padded (Section~\ref{Overview}). 
The transformer block operates on the slice according to 
\begin{equation}
\mathbf{Z}_{k,j} = TB(\mathbf{E}_{k,j}). 
\label{eq:localblock}
\end{equation}
After applying $TB$, the zero-padded feature vectors are reset to zero according  to the binary attention masks. 
The output of the local block, $\mathbf{Z}_j^{loc}$, which collects the cluster-specific slices $\mathbf{Z}_{k,j}$, preserves the input dimensionality, i.e.\ its dimensions are $K \times (S_c + 1) \times N_F$, but it encodes enriched local representations. 


This mechanism captures intra-cluster dependencies and aggregates local geometric context among faces. 
Each local block independently processes the feature sequences within a cluster, enabling every feature vector to attend to and influence others within its local region. 
The attention-based interaction also applies to the cluster token. 
On one hand, each face feature vector is influenced by the cluster token of its cluster; on the other hand, the cluster token aggregates information from all face features in the cluster. 
This bidirectional exchange serves two key purposes: (1) the cluster token captures a summary representation of its cluster, and (2) this aggregated information is redistributed to all member features. 
As described in Section~\ref{sec:globalblock}, in the cross-cluster block, cluster tokens from all clusters interact, so that in subsequent local blocks, the cluster tokens also encode global context that influences all face-level features.


\subsubsection{Cross-Cluster Block: }
\label{sec:globalblock}

This block addresses dependencies between clusters and thus captures global context with minimal computational overhead. 
Only the cluster tokens participate in this process.
First, the cluster tokens are extracted from the output of the local block, $\mathbf{Z}_j^{loc}$, and combined to a tensor  $\mathbf{Z}_{j}^{cl,in}$ of dimensionality $K \times N_F$, which serves as the input to another transformer block of the same structure as the one described in Section \ref{sec:localblock}. 
This $TB$ generates an output tensor of the same dimension as the input:
\begin{equation}
\mathbf{Z}_{j}^{cl,out} = TB(\mathbf{Z}_{j}^{cl,in}).
\label{eq:globalblock}
\end{equation}
The transformed cluster tokens are then copied back into the output tensor of the local block $\mathbf{Z}_j^{loc}$, producing the final output $\mathbf{Z}_j$ of TSTB $j$. 
This output forms the input for the next TSTB, i.e.\ 
$\mathbf{E}_{j+1} = \mathbf{Z}_j$. 
Thus, the updated cluster tokens propagate global information into the subsequent block. 
The output $\mathbf{Z}_{N_{Bl}}$ of the final TSTB block serves as the input to the classification head described in Section~\ref{sec:classification_head}.


\subsection{Classification Head }
\label{sec:classification_head}
The classification head uses the output $\mathbf{Z}_{N_{Bl}}$ of the last TSTB  to predict one class label for every face of the input mesh based on the feature vector representing that face, which has accumulated local and global information. 
Before doing so, the cluster tokens, the zero-padded dummy face tokens  and the arrangement of the faces according to the clusters, still maintained in the structure of $\mathbf{Z}_{N_{Bl}}$, are discarded. 
Thus, a new tensor $\mathbf{Z}_F$ of dimension $\lvert F \rvert \times N_F$ is generated. 
Then, each of these $N_F$-dimensional feature vectors is transformed to a vector $\mathbf{y}_i$ containing $C$ raw class scores using a fully connected layer. 
Finally, these raw class scores are normalized using the softmax function, yielding normalized class scores for face $i$, 
\begin{equation}
\hat{\mathbf{y}}_i = \mathrm{softmax}(\mathbf{y}_i)\, ,
\end{equation}
collected in the tensor $\widehat{Y}$ introduced in Section~\ref{Overview}. 
The final prediction for each face is obtained as the class label having the maximum  class score.


\subsection{Training}
\label{sec:loss}
The network is trained by minimizing a categorical cross-entropy loss $L_{\text{ce}}$, which measures the discrepancy between the predicted class probabilities for each mesh face and the reference labels:
\begin{equation}
L_{ce}= -\frac{1}{M} \sum_{i \in \mathcal{B}} \sum_{c=1}^{C} y_{ic} \log(\hat{y}_{ic}).
\label{eq:minibatch_loss}
\end{equation}
In every iteration, Equation~\ref{eq:minibatch_loss} forms the basis of an update step, considering a minibatch of the training samples.  
The index set of the sampled faces in the current minibatch is denoted by $\mathcal{B} \subseteq \{1, \dots, |F|\}$, with $|\mathcal{B}| = M$. 
$M$ can differ from the number of faces $|F|$, because some faces may be unlabelled. 
The binary indicator $y_{ic}$ specifies whether face $i$ belongs to class $c$ ($y_{ic} = 1$) or not ($y_{ic} = 0$), and $\hat{y}_{ic}$ denotes the predicted probability for face $i$ to be assigned to class $c$. 
Details about the training procedure used in the experiments can be found in  Section~\ref{Setup}. 


\section{Experiments and Results} 


\subsection{Datasets}
\label{Datasets}

Two datasets were used for evaluating our method: SUM, a benchmark for urban semantic segmentation, and a new dataset designed for detecting damages in a cultural heritage site (CH). 


\subsubsection{Semantic Urban Meshes (SUM):} 
This dataset,  covering an urban area of approximately 4~km$^2$ in  the centre of Helsinki (Finland),  was introduced by \citet{gao2021sum}.  
It consists of textured 3D meshes reconstructed from airborne oblique imagery. 
The average side length of a triangle is approximately  0.8~m, and the ground sampling distance (GSD) of the texture is  10~cm. 
The faces are annotated with one of six semantic categories typical for urban classification ({\em terrain} - {\em T}, {\em high vegetation} - {\em V}, {\em building} - {\em Bld}, {\em water} - {\em W}, {\em car} - {\em C}, and {\em boat} - {\em B}), resulting in approximately 19 million labelled  faces. 
The data is split into 64 tiles of about 250~m$^2$ each, with 40 / 12 / 12 of them to be used for training, validation and testing, respectively. 


\subsubsection{Damage Detection in Cultural Heritage (CH):} This dataset was designed for evaluating methods  for  damage detection in CH assets. 
It was generated in  the context of the  project ChemiNova, in which the development of the proposed method is embedded.  
It is based on a textured 3D mesh of a part of the roof of the La Nau  Centre in Valencia (Spain). 
The mesh was generated from 428 aerial RGB images acquired by the built-in camera of a DJI Mini 4 Pro drone with a focal length of 7~mm. 
The distance from the object was about 5~m, and the images were acquired with 80\% forward  and 70\% side laps. 
Agisoft Metashape\footnote{https://www.agisoft.com/ (accessed 13/02/2026)} was used to orient the images using Structure-from-Motion and to generate a dense 3D point cloud as well as the textured 3D mesh. 
The positions of the drone determined by the built-in GNSS sensor were used for georeferencing the block in a global reference frame (WGS84 / UTM 30N). 
The average side length of a triangle in the mesh is about 7~mm, the GSD of the texture is  1~mm.


To generate a reference, expert conservators defined a class structure of damages occurring on the roof that consists of seven categories. 
Two of them encode different states of undamaged areas: {\em No Damage} ({\em ND}) and {\em Joint} ({\em J}), corresponding to bright and dark undamaged areas, respectively. 
The remaining five classes encode actual damages: {\em Stain} ({\em St}), {\em Deposit} ({\em D}), {\em Loss of Structural Integrity} ({\em LI}), {\em Biological Colonization} ({\em BC}) and {\em Salt Efflorescence} ({\em SE}). 
A part of the 3D mesh was annotated by the conservation experts by  digitizing polygons in an orthophoto of the roof and assigning them to the classes just mentioned; the results were  projected onto the mesh to obtain precise triangle-wise annotations. 
Triangles without annotations were marked as being {\em unlabelled}. 
Finally, the annotated mesh was spatially partitioned into 230 tiles containing about 30k–50k triangles each, which were divided into training, validation, and test subsets. 
Table~\ref{tab:ch_dataset_distribution} presents the class distribution of the triangles  and the assignment of tiles to the training, validation and test datasets. 
The class distribution is very imbalanced, with the two classes encoding undamaged areas corresponding to about 70\% of the data. 
The dominating damage type is {\em Biological Colonization} (18\% of the samples), whereas the damage types {\em Deposit} and {\em  Salt Efflorescence} are extremely underrepresented. 
In total, the CH dataset contains about 5 million annotated triangles. 


\begin{table}[!ht]
\centering
\begin{tabular}{lrrrr}
\toprule
{\em Class} & {\em Training} & {\em Validation} & {\em Testing} & {\em Total} \\
\midrule
{\em ND}   & 1,445,547 & 348,638 & 530,823 & 2,325,008 \\
{\em J}  & 778,955   & 182,090 & 234,293 & 1,195,338 \\
{\em St}  & 337,499   & 48,771  & 119,975 & 506,245 \\
{\em D}  & 8,185     & 1,342   & 2,837   & 12,364 \\
{\em LI}   & 67,742    & 19,573  & 22,448  & 109,763 \\
{\em BC}  & 575,191   & 144,778 & 202,052 & 922,021 \\
{\em SE}  & 15,107    & 6,233   & 5,265   & 26,605 \\
\midrule
{\em Total} &3,228,226 & 751,425 &  1,117,693 & 5,097,344 \\
\midrule
{\em Tiles} & 133 &29 & 41 & 203 \\
\bottomrule
\end{tabular}
\caption{Number of triangles per class and number of tiles in the CH dataset and the training, validation and test subsets.}
\label{tab:ch_dataset_distribution}
\end{table}


\subsection{Experimental Setup}
\label{Setup}


\subsubsection{Hyperparameters Training: }
\label{hyperparameters}
All experiments were conducted on a single NVIDIA A100 GPU (40 GB memory). 
Network hyperparameters were determined empirically based on validation set performance. 
The same hyperparameter values were used for both datasets, except for ablation studies. 
The  dimensionality  of the geometrical and texture feature vectors (Section~\ref{Feature_Extraction}) was set to $d = 64$. 
While transformers can, in principle, process sequences of arbitrary length, a uniform input length is required in practice. Therefore, we fix the number of pixels per face to $P = 128$ in the texture branch (Section~\ref{Texture Branch}). 
Faces containing more than 128 pixels are truncated (this occurs for fewer than 2\% of samples in both datasets), whereas smaller faces are zero-padded to this length. 
Similarly to what is done in the TSTB (Section~\ref{Overview}), an attention mask is used to ignore padded tokens during processing. 
Beyond the feature extraction branch, the network  comprises $N_{Bl} = 6$ TSTBs, each with two attention heads and an embedding dimension of $N_F = 256$. 
The number of clusters (Section~\ref{Overview}) was set to $K = 300$.


The network was trained on tiles from the training subsets of the datasets by minimizing the objective defined in Section~\ref{sec:loss}. 
Model weights were initialized randomly according to  \citep{glorot2010understanding}. 
We used the AdamW optimizer \citep{loshchilov2017decoupled} with an initial learning rate of $\eta = 1 \times 10^{-4}$. 
A cosine annealing schedule was used to progressively decay the learning rate to $1 \times 10^{-6}$ over the first half of training epochs, following the strategy of~\citet{loshchilov2017decoupled}.
To mitigate overfitting, we applied extensive data augmentation, including random rotations of up to $\pm 45^\circ$ about all three axes, random scaling with factors in the range $[0.5, 2]$, and additive Gaussian noise ($\sigma = 0.01$) applied to vertex positions to improve robustness to geometric perturbations. 
The final model weights are selected based on the highest $mF1$ score achieved on the validation set during training.


\subsubsection{Evaluation Metrics: }
\label{eval}

Performance was evaluated by comparing the labels predicted for the  test sets to the provided reference. 
We determined the class-specific $F1$ scores and  the mean F1 score over all classes ($mF1$).  
For most experiments we only report $mF1$ and the overall accuracy ($OA$), the latter corresponding to the percentage of correctly classified faces. 


\subsubsection{Ablations: }
\label{setup_ablations}

We also performed several ablation studies. 
First, we assess the impact of the embedding dimensionality $N_F$ on the result (Section~\ref{Feature_Extraction}). 
Second, we analyze the results for different values of the number $N_{Bl}$ of TSTBs in our model (Section~\ref{Overview}). 
Third, we analyze the  contribution of the input modalities on the classification results. 
This is achieved by training two additional classifiers that either only use the geometrical features $\mathbf{F}_G$ or only the texture features $\mathbf{F}_T$. 
These first three ablations were only performed on the SUM dataset. 
A fourth ablation is related to the impact of the way in which the features are processed in the TSTBs. 
In particular, we compare our method to another one that uses the cross-attention block of \citet{heidarianbaei2025nomeformer} instead of our cross-cluster block in the TSTB. 
This comparison, which is supposed to highlight the advantages of the new architecture of that block, is performed on both test datasets. 


\subsubsection{Baseline Methods: }
\label{setup_baselines}

As the manifoldness constraint is not satisfied in either of our datasets, we had to select baselines capable of handling non-manifold  meshes. 
Among the few existing methods (Section~\ref{sec:Related_Work}), we
selected DiffusionNet \citep{sharp2022diffusionnet} and NoMeFormer \citep{heidarianbaei2025nomeformer} as baselines, in both cases relying on the original hyperparameters reported in the respective papers. 
It is worth noting that DiffusionNet completely disregards textural information, while NoMeFormer relies solely on pixel statistics. 
The comparison to these baselines is based on the performance metrics presented in Section~\ref{eval}. 
In addition, for the SUM benchmark, we compare our results to those of the best-performing methods on that dataset according to \citep{gao2021sum}:  RF-MRF \citep{rouhani2017semantic} and KPConv \citep{thomas2019kpconv}. 
For that comparison, we had to adapt the quality metrics: \citet{gao2021sum} weight each triangle by its area when computing $OA$ and $mF1$, so we present these scores according  to the same definition. 
KPConv is a point-cloud based method, which was applied to classify points sampled from the triangulated mesh. 
In this case, it is not exactly clear how the quality indices were determined; we use the numbers presented in \citep{gao2021sum}. 


\subsection{Results}
\label{results}


\subsubsection{Results on the SUM Dataset: }
\label{results:SUM}

Table~\ref{tab:f1_scores_SUM} summarizes the results of the evaluation of our method  on the SUM dataset. 
The $OA$ of 94.3\% is rather satisfactory, indicating a good performance of our method: less than 6\% of the class labels are wrong. 
The  $mF1$ score is somewhat lower at 81.9\%, which still seems to be acceptable. 
In particular, for the classes {\em terrain}, {\em high vegetation} and {\em building}, the F1 scores are larger than  92\%, which we consider very good. 
However, the difference between $mF1$ and $OA$ indicates some problems with less frequent classes.
For {\em water}, {\em car} and {\em boat},  the F1 scores drop to 78.8\%, 62.9\%, and 64.3\%, respectively. 
In particular,  {\em car} and {\em boat}  exhibit a high intra-class variability. 
With only a handful of training samples available for these classes, achieving strong generalization is inherently challenging.


\begin{table}[!ht]
\centering
\begin{tabular}{lccccccc}
\hline
{Class} &  {\em T} &  {\em V} & {\em Bld} &  {\em W} & {\em C} & {\em B}  \\ 
\hline
{$F1$ [\%]} & 92.3 & 94.5 & 95.6 & 78.8 & 62.9 & 64.2   \\ 
\hline
\end{tabular}%
\caption{Class-specific F1 scores of our method achieved on the SUM dataset. The $mF1$  score is 81.9\%, the $OA$ is 94.3\%.}
\label{tab:f1_scores_SUM}
\end{table}


\subsubsection{Results on the CH Dataset: }
\label{results:CH}
Table~\ref{tab:f1_scores} presents the evaluation of the results of our method achieved on the CH dataset. 
The $OA$ of 72.8\% seems to be acceptable, though it is certainly lower than the one achieved on the SUM dataset. 
At 49.7\% the $mF1$ score is considerably lower, with the difference to $OA$ again indicating problems with underrepresented classes. 
Notably, the two classes corresponding to undamaged triangles ($ND$ and $J$), which together account for approximately 70\% of the dataset, achieve F1 scores of about 80\%, indicating a relatively strong performance in the dominant categories. 
In contrast, the classes $D$ and $SE$ have the lowest F1 scores (0.0\% and 11.7\%, respectively). 
This performance drop is positively correlated with the number of samples in these categories, highlighting the severe class imbalance and underrepresentation that limit the model’s ability to learn minority class features. 
The damage classes with the highest F1 scores are $LI$ and $BC$. 
The latter represents the type of damage occurring most frequently in the data. 
The relatively strong performance for $LI$, for which the number of samples is not very large, is more surprising; perhaps this damage type leads to specific geometrical features. 
These results indicate the potential of the method for such challenging datasets, but also highlight its limitations. 


\begin{table}[!ht]
\centering
\begin{tabular}{lccccccc}
\hline
{Class} & $ND$ & $J$& $St$ & $D$ & $LI$ & $BC$& $SE$   \\ 
\hline
{$F1$ [\%]} & 78.3 & 81.8 & 50.1 & 0.0 & 62.9 & 63.4 & 11.7   \\ 
\hline
\end{tabular}%
\caption{Class-specific F1 scores of our method achieved on the CH dataset. The $mF1$ score is 49.7\%, the $OA$ is 72.8\%.}
\label{tab:f1_scores}
\end{table}


\subsubsection{Ablation Studies: }
\label{results:ablations}

Table~\ref{tab:embedding_dim_vs_mf1} shows the $mF1$ scores and overall accuracies achieved on the SUM test set when using our method with  different values of the feature dimension~$N_F$. 
Increasing $N_F$ from 32 to 256 improved the $mF1$ scores and $OA$ values by  7\% and 3.5\%, respectively, which shows that this hyperparameter has a significant impact on the results. 
The table indicates that using larger values might still lead to an increase in the performance, but this could not be tested due to hardware constraints.


\begin{table}[h!]
\centering
\begin{tabular}{rcc}
\toprule
${N_F}$ &${OA}$  [\%]& ${mF1}$ [\%] \\
\midrule
32 &90.6& 74.9 \\
64 &90.8& 75.4 \\
128& 92.9 & 78.6  \\
256& 94.3 & \textbf{81.9} \\
\bottomrule
\end{tabular}
\caption{Effect of the feature dimension ${N_F}$ on model performance on the SUM dataset. }
\label{tab:embedding_dim_vs_mf1}
\end{table}

A comparison of the results achieved on the SUM test set by using different numbers $N_{Bl}$ of TSTBs showed that using only two such blocks resulted in a suboptimal performance of 78.5\% in $mF1$, which compares to a $mF1$ score of  81.9\%  achieved with $N_{Bl}=6$ blocks, our default configuration. 
Further increasing the number of blocks to $N_{Bl}=8$ resulted in an insignificant improvement (smaller than  0.05\%).

Table~\ref{tab:modality} reports the  performance metrics achieved by our method when using different input modalities. 
Using geometry alone yields the lowest $mF1$ score and $OA$ of 66.9\% and 84.1\%, respectively, indicating that geometric cues  are insufficient for reliable classification on the SUM dataset. 
Using texture alone results in an improvement of $mF1$ by 3.3\% ($mF1=70.2\%$). 
This  suggests that texture carries more discriminative information. 
However, combining both geometry and texture achieves a substantially higher $mF1$ score and $OA$ of 81.9\% and 94.3\%, respectively, demonstrating the complementary nature of the two modalities and the advantages of their joint usage.


\begin{table}[h!]
\centering
\begin{tabular}{lcc}
\toprule
Modality &${OA}$ [\%]& ${mF1}$ [\%]  \\
\midrule
Geometry &84.1 &66.9 \\
Texture &88.1 &70.2 \\
Both  & 94.3 &\textbf{81.9} \\
\bottomrule
\end{tabular}
\caption{Comparison of quality metrics achieved on the SUM dataset when using different input modalities.}
\label{tab:modality}
\end{table}


Finally, we compare two different strategies for modeling long-range interactions. 
Table~\ref{tab:attention} shows that our cross-cluster blocks based on self-attention between cluster tokens (S-A) outperforms the cross-attention based approach of \citet{heidarianbaei2025nomeformer} (C-A). 
The difference is relatively small on the SUM dataset (0.7\% in $mF1$, 0.5\% in $OA$); it is more pronounced on the CH dataset, with an increase of 3.4\% in $mF1$ and of 2.8\% in $OA$. 
We hypothesize that this difference between the datasets is related to  the object size in relation to the resolution  of the data. 
Some of the damage categories in the CH dataset occur in spatially small clusters that might vanish  in the cross-attention scenario, where the face feature vectors are updated by weighted averages of all cluster features. 
In contrast, the SUM dataset consists of larger, urban-scale structures for which the impact of the global interactions might be limited.


\begin{table}[h!]
\centering
\begin{tabular}{lcccc}
\toprule
\multirow{2}{*}{Method} & \multicolumn{2}{c}{{SUM}} & \multicolumn{2}{c}{{CH}} \\
\cmidrule(lr){2-3} \cmidrule(lr){4-5}
 & ${OA}$& ${mF1}$  & ${OA}$ & ${mF1}$  \\
  & [\%]&  [\%]  &  [\%]&  [\%] \\
\midrule
C-A & 93.8 & 81.2 & 70.0 & 46.3 \\
S-A (ours) & \textbf{94.3} & \textbf{81.9} & \textbf{72.8} & \textbf{49.7} \\
\bottomrule
\end{tabular}%
\caption{Comparison of the results achieved on the two test datasets using two architectures of the TSTBs. C-A: global interactions are considered by cross-attention, following \citet{heidarianbaei2025nomeformer}. S-A: our proposed method using self-attention in the cross-cluster blocks.}
\label{tab:attention}
\end{table}


\subsubsection{Comparison to other Methods: }
\label{results:comparison}

\begin{figure}[!ht]
    \centering
    \includegraphics[width=0.48\textwidth]{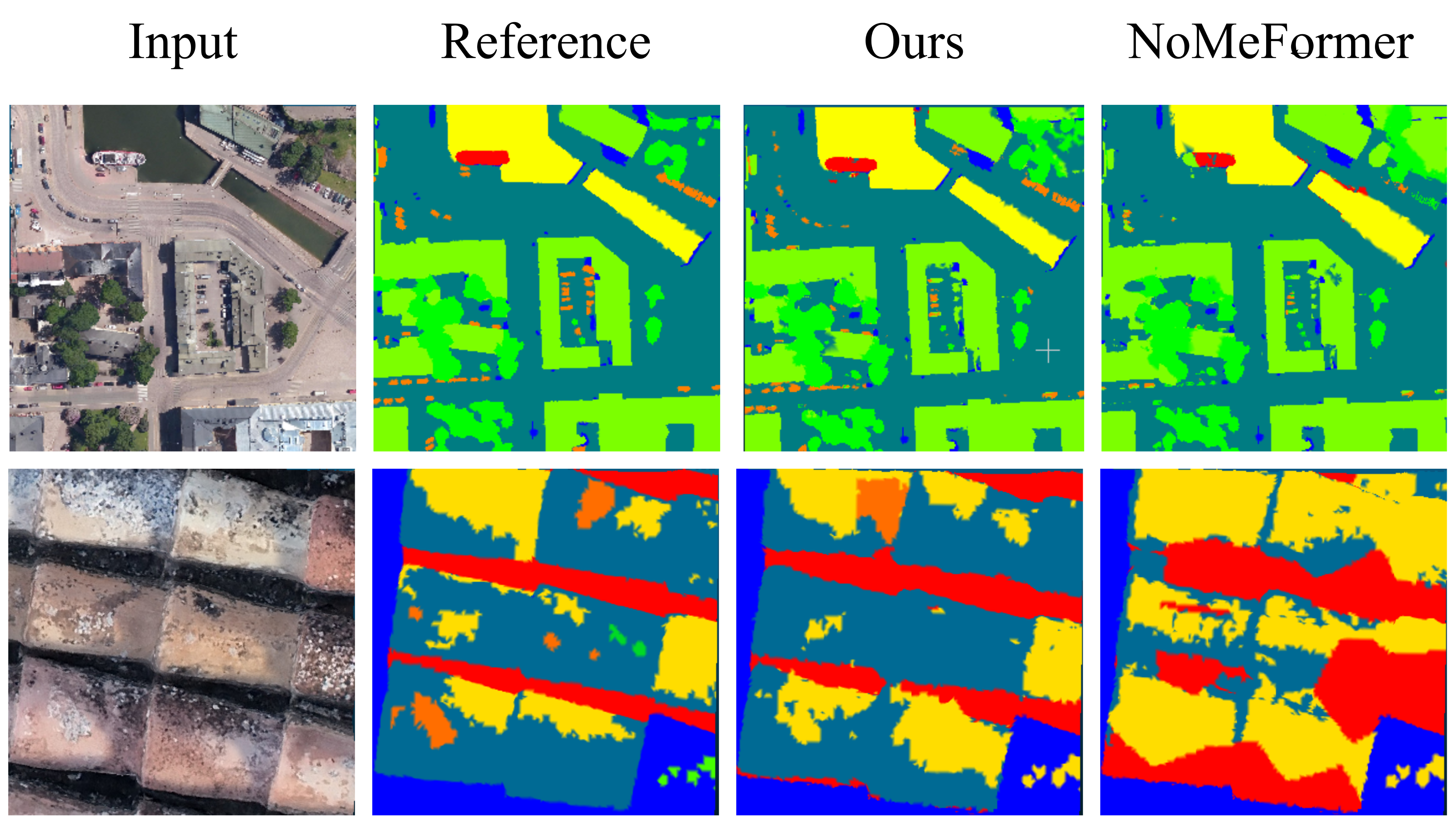}
    \caption{Qualitative results achieved by our method and by the best-performing baseline, NoMeFormer. From left to right: Input, reference, our method, and  NoMeFormer. Top: an example from SUM. Bottom: an example from CH. Colour code — SUM: $T$ (blue), $V$ (dark green), $Bld$ (green), $W$ (yellow), $C$ (orange), $B$ (red). Colour code — CH: $ND$ (greenish blue), $J$ (red), $St$ (green), $D$ (chartreuse green), $LI$ (gold), $BC$ (yellow), $SE$ (orange), unlabelled (blue).
    }
    \label{fig:quantative}
\end{figure}

Table~\ref{tab:Performance} presents the comparison of the results achieved by our method to the two base\-lines presented in Section~\ref{setup_baselines}. 
Figure~\ref{fig:quantative} shows some qualitative results achieved by our method and the best-performing baseline. 
Our method outperforms these baselines by a large margin in all evaluation metrics. 
Compared to the best-per\-forming baseline, NoMeFormer, the improvement in $mF1$ and $OA$ is 15.3\% and 11.7\%, respectively, on the SUM dataset. 
On the CH dataset, the improvement in $mF1$ and $OA$ achieved by our method over NoMeFormer is 15.6\% and 4.9\%, respectively. 
The results for DiffusionNet are still worse than those of NoMeFormer, with  differences in $mF1$ larger than 20\% on both datasets. 
We attribute the better performance of our method partly to its ability to leverage the rich textural information  that cannot be fully captured through statistical texture descriptors alone. 
Figure~\ref{fig:quantative} indicates that compared to NoMeFormer, our model produces notably sharper and more accurate predictions across all semantic classes. 
The results of NoMeFormer show a blurring effect, which we attribute to the properties of the global aggregation module used in \citep{heidarianbaei2025nomeformer}. 
The integration of cross-cluster blocks mitigates this problem,  leading to a better balance between local and global features, although some fine-scale structures remain partially unresolved.

\renewcommand{\arraystretch}{1.0}
\begin{table}[h!]
\centering
\begin{tabular}{lcccc}
\toprule
\multirow{2}{*}{Method} & \multicolumn{2}{c}{SUM} & \multicolumn{2}{c}{{CH}} \\
\cmidrule(lr){2-3} \cmidrule(lr){4-5}
 & $OA$ & $mF1$ & $OA$ & $mF1$ \\
\midrule
DiffusionNet & 80.4 & 60.1 & 63.8 & 25.9 \\
NoMeFormer   & 84.3 & 66.6 & 67.9 & 34.1 \\
\textbf{Ours} & \textbf{94.3} & \textbf{81.9} & \textbf{72.8} & \textbf{49.7} \\
\bottomrule
\end{tabular}%
\caption{Overall Accuracy ($OA$) and mean F1 score ($mF1$), both in [\%], achieved by our method and two baselines on the SUM and CH datasets.}
\label{tab:Performance}
\end{table}


Table~\ref{tab:Performance_wighted}  compares the results of different methods  on the SUM dataset under the area-weighted evaluation protocol of \citep{gao2021sum}, which prevents a disproportionate influence of densely sampled regions  in datasets with large variations in triangle size. 
The results of DiffusionNet and NoMeFormer were achieved in our own experiments  while those of the other methods are taken from \citep{gao2021sum}. 


\renewcommand{\arraystretch}{1.0}
\begin{table}[h!]
\centering
\begin{tabular}{lcc}
\toprule
Method & $OA$ & $mF1$ \\
\midrule
RF-MRF       & 91.2 & 68.1 \\
KPConv       & 93.3 & 76.7 \\
\midrule
DiffusionNet & 81.0 & 61.3 \\
NoMeFormer   & 84.9 & 68.6 \\
\textbf{Ours} & \textbf{93.3} & \textbf{78.4} \\
\bottomrule
\end{tabular}%
\caption{Area-weighted $OA$ and $mF1$ scores [\%] achieved by our method and four baselines on the SUM dataset.}
\label{tab:Performance_wighted}
\end{table}


As shown in the table, our method achieves the highest performance with a $mF1$ score of 78.4\%, despite a marginal drop of 3.5\% compared to the unweighted metric shown in Table~\ref{tab:Performance}. 
The fact that for NoMeFormer and DiffusionNet the area-weighted metrics are slightly better than the unweighted ones  suggests that these methods are more effective on larger mesh faces, where local geometric context dominates. 
In any case, the performance gain of our method compared to these two baselines is still obvious. 
Compared to the best-performing methods according to \citep{gao2021sum}, our method also performs better. 
It achieves the same $OA$ as the point-based KPConv, but achieves a better area-weighted $mF1$ score (+1.7\%), indicating the effectiveness of directly operating on mesh faces rather than relying on a point-cloud approximation, although the exact experimental protocol used in \citep{gao2021sum} is unclear. 
Compared to the mesh-processing framework RF-MRF, the area-weighted $mF1$ score achieved by our method is 10.3\% higher; the improvement in $OA$ is less pronounced, indicating even stronger problems for RF-MRF with underrepresented classes. 
The performance margin compared to RF-MRF indicates the advantage of deep learning for mesh-structured data.


\section{Conclusion and Future Work}
This work introduces a new transformer-based architecture for semantic segmentation of textured 3D meshes. 
In contrast to prior approaches that represent texture information by low-di\-mensional statistics, our model directly encodes texture through a transformer-based texture extraction branch, thereby preserv\-ing fine-grained appearance cues that are critical for distinguishing geometrically similar classes. 
Furthermore, a new approach to short- and long-range representation learning is proposed: the local mesh structure is captured through face-level processing, while global context is modelled via self-attention over cluster tokens, allowing to consider long-range interactions. 


Extensive experiments performed on two datasets from different domains, including a new dataset for damage detection on a CH building, have demonstrated consistent performance gains of our method compared to existing ones. 
On the SUM benchmark, our model achieved  a $mF1$ score of 81.9\% and an $OA$ of 94.3\%, surpassing mesh- and point-based baselines alike. 
On the CH dataset, our method obtained a $mF1$ score of 49.7\% and an $OA$ of 72.8\%, markedly improving over prior non-manifold mesh transformers. 
Ablations verify that replacing global cross-attention with cluster-token self-attention preserves fine-scale cues and improves performance (\(+0.7\%\)  on SUM and \(+3.4\%\) on CH in $mF1$). 
These ablation studies  confirm that incorporating raw texture information and enhancing global communication greatly improve mesh-based semantic segmentation.


Despite the promising results, several directions remain open. 
The most notable limitation of our method is the absence of explicit positional encoding in the texture transformer branch, which we leave for future exploration. 
In addition, memory and runtime constraints currently limit the number of faces and texture features that can be processed simultaneously, highlighting the need for more scalable attention mechanisms and model compression techniques. 
Additional performance gains may be achieved by integrating end-to-end learnable clustering, loss re-weighting for class imbalance, and leveraging self-supervised pre-training on joint mesh geometry and texture. 
Extensions of additional modalities, including multispectral or thermal data, also represent compelling avenues for future development. 


\section*{Acknowledgments}
The research reported in this paper was performed in the context of the project ChemiNova (https://cheminova.eu/). ChemiNova has received funding from the European Union’s Horizon Europe Framework Programme under grant agreement 101132442.


\bibliography{main}
\end{document}